# Learning Subgroup Relations Using Siamese Graph Neural Networks

Tal Weissblat

Ramat Gan Academic College, Ramat Gan 52275, Israel

Correspondence: tal.weisblat@gmail.com
ORCID: 0009-0001-9160-6790

## Abstract

Determining whether one finite group is isomorphic to a subgroup of another is a fundamental problem in computational group theory. In this work, we propose a Siamese Graph Neural Network (Siamese GNN) for subgroup prediction using Cayley graph representations of finite groups. Each input group is represented by its undirected Cayley graph and encoded by one branch of a Siamese GNN to produce a graph embedding. The resulting graph embeddings are combined with algebraic features derived directly from the input groups to construct a joint feature vector, which is processed by a fully connected classifier to predict subgroup relations between finite groups. By integrating graph-based structural representations with algebraic features, the proposed framework provides a unified approach for learning subgroup relations from finite groups. Experimental results on an expanded and more diverse dataset of 308 finite-group pairs drawn from 11 group families demonstrate the effectiveness of the proposed architecture, achieving a test BA of 91.67% on an independent test set. Additional experiments evaluate generalization to unseen groups, robustness to different Cayley graph generating sets, the contribution of GNN message passing, performance relative to non-neural baselines, and comparison with exact computational methods. These results illustrate the potential of geometric deep learning for subgroup prediction.

## 1 Introduction

Determining whether one finite group is isomorphic to a subgroup of another is a fundamental problem in computational group theory. Subgroup structure plays a central role in the study and classification of finite groups, with applications in computational algebra, symmetry analysis, and many other areas of mathematics. Classical computational approaches employ exact algebraic algorithms to analyze finite groups and their subgroup structures [1]. Although these methods are mathematically rigorous and highly effective, recent advances in machine learning have introduced new opportunities for learning algebraic relationships directly from suitable representations of groups, complementing traditional symbolic techniques.

The objective of the proposed approach is not to replace exact computational group-theoretic algorithms, which remain the appropriate tools when an exact determination of subgroup relations is required. Rather, this work investigates whether algebraic relationships between finite groups can be learned from their graph representations. In particular, the study examines whether Cayley graphs encode structural information that can be captured by GNNs and used to learn subgroup relations. The proposed framework should therefore be viewed as a learning-based investigation of graph representations of algebraic structures, complementary to exact symbolic methods.

One particularly natural representation of a finite group is its Cayley graph, which encodes the multiplication structure of the group with respect to a generating set [2]. Cayley graphs provide a graph-theoretic representation of algebraic objects while preserving important structural information, making them a natural bridge between computational group theory and graph-based machine learning.

At the same time, Graph Neural Networks (GNNs) have emerged as successful deep learning frameworks for learning from graph-structured data [3,4]. Early GNN models established neural architectures capable of operating directly on graph-structured information [5]. Subsequent developments introduced graph convolutional approaches that learn representations by combining graph structure with node features [6], as well as inductive neighborhood-aggregation methods that generate representations by aggregating information from local neighborhoods [7]. Through repeated neighborhood aggregation, GNNs can capture structural information from increasingly larger neighborhoods and construct representations suitable for node- and graph-level learning tasks [4,8]. More recent theoretical studies have investigated the expressive power of GNNs, characterizing their ability and limitations in distinguishing different graph structures [8].

Motivated by these developments, recent studies have explored the use of GNNs for learning algebraic properties directly from Cayley graph representations. In particular, previous studies demonstrated that GNNs can successfully predict important algebraic properties of finite groups directly from their Cayley graphs [9,10]. These findings suggest that Cayley graphs constitute an informative representation for combining computational group theory with geometric deep learning.

Building on these previous studies, the present work considers a different but closely related learning problem. Whereas previous studies focused on predicting properties of individual groups, we investigate whether one finite group is isomorphic to a subgroup of another. This pairwise prediction task requires simultaneously processing two groups and learning the structural relationship between them. Related GNN-based approaches have demonstrated that neural representations can be learned for comparing pairs of graph-structured objects [11]. Consequently, the proposed framework adopts a Siamese architecture [12,13], in which two identical GNN encoders with shared parameters independently process the corresponding Cayley graphs. Sharing the encoder ensures that both groups are represented in the same embedding space, enabling their structural properties to be compared consistently. The resulting graph embeddings are combined through graph-derived features and integrated with algebraic features derived directly from the input groups to form a joint feature vector, which is then processed by a fully connected classifier to predict whether one group is isomorphic to a subgroup of the other. This modular architecture separates representation learning, performed by the shared GNN encoder, from decision making, performed by the classifier, allowing either component to be replaced or modified independently. By integrating graph-based structural representations with algebraic features, the proposed framework provides a unified approach for learning subgroup relations from finite group representations.

The main contributions of this work are threefold. First, we propose a Siamese GNN for predicting subgroup relations between finite groups using their Cayley graph representations. Second, we integrate graph-level embeddings learned from Cayley graphs with algebraic features derived directly from the input groups, enabling the model to exploit both structural and algebraic information. Third, we conduct a comprehensive experimental evaluation, including different data-split setups, cross-validation, an ablation study, alternative Cayley graph generating sets, non-neural baselines, and comparison with an exact computational approach, demonstrating the effectiveness of the proposed framework for subgroup prediction.

## 2 Methodology

This section presents the proposed methodology for subgroup prediction using a Siamese GNN. As illustrated in Figure 1, each input group is first represented as an undirected Cayley graph. The two graphs are then processed by a shared GNN encoder to obtain graph embeddings. In parallel, algebraic features are extracted directly from the input groups. The graph-derived features obtained from the embeddings and algebraic features are combined into a joint feature vector, which is then processed by a classifier to predict whether $H$ is isomorphic to a subgroup of $G$. The following subsections describe each stage of the proposed framework in detail.

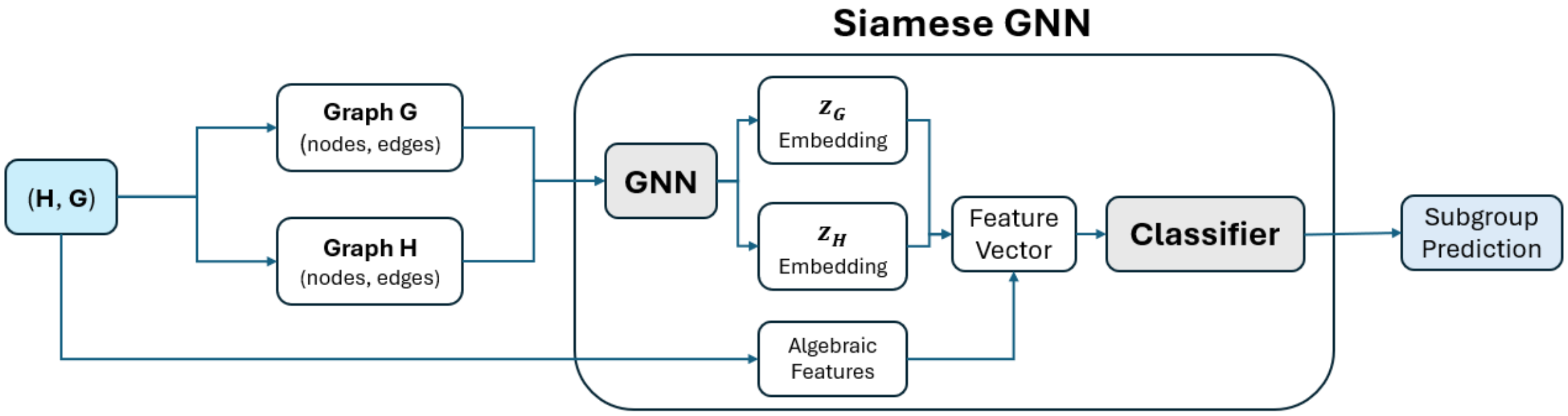

**Figure 1.** Overview of the proposed methodology. The input groups are represented as undirected Cayley graphs and encoded by a shared GNN encoder to produce graph embeddings. In parallel, algebraic features are extracted directly from the input groups. The graph-derived features obtained from the embeddings and algebraic features are combined into a joint feature vector, which is processed by a classifier to predict whether $H$ is isomorphic to a subgroup of $G$.

### 2.1 Dataset

The dataset consists of 308 pairs of finite groups, with each pair assigned a binary label indicating whether the first group is isomorphic to a subgroup of the second. The groups represented in the dataset are drawn from 11 families: cyclic, dihedral, symmetric, alternating, quaternion, special linear, finite Heisenberg, unitriangular, general linear, affine general linear, and projective special linear groups.

To evaluate the generalization ability of the proposed model under progressively stricter conditions, three data-split setups are considered: Standard, Test-group-disjoint, and Fully group-disjoint. In the Standard split, group pairs are partitioned into training, validation, and test sets without imposing restrictions on the occurrence of individual groups across the three partitions; consequently, the same group may appear in more than one partition. In the Test-group-disjoint split, all groups appearing in the test set are excluded from both the training and validation sets, while groups may still be shared between the training and validation sets. Finally, in the Fully group-disjoint split, the training, validation, and test sets contain mutually disjoint groups, such that no group appearing in one partition appears in either of the other two.

The additional disjointness constraints reduce the number of group pairs that can be retained in the corresponding experimental setups. In particular, when moving from the Standard split to the Test-group-disjoint split, pairs that would violate the separation of the test groups from the training and validation groups are excluded. The Fully group-disjoint split imposes the strictest condition, requiring complete group separation across all three partitions and consequently excluding additional pairs that cannot be assigned without violating this constraint. Thus, the total number of retained pairs may decrease as increasingly strict group-disjoint conditions are imposed. Table 1 summarizes the numbers of positive and negative pairs in the training, validation, and test partitions for each of the three data-split setups.

**Table 1.** Dataset statistics under the three data-split setups.

| Split setup | Set | Neg. | Pos. | Total | % Neg. | % Pos. | % Total |
|---|---|---|---|---|---|---|---|
| Standard split | Train | 100 | 54 | 154 | 64.94% | 35.06% | 50.00% |
| | Validation | 30 | 31 | 61 | 49.18% | 50.82% | 19.81% |
| | Test | 54 | 39 | 93 | 58.06% | 41.94% | 30.19% |
| | **Total** | **184** | **124** | **308** | 59.74% | 40.26% | 100.00% |
| | | | | | | | |
| Test-group-disjoint | Train | 58 | 48 | 106 | 54.72% | 45.28% | 51.46% |
| | Validation | 54 | 27 | 81 | 66.67% | 33.33% | 39.32% |
| | Test | 12 | 7 | 19 | 63.16% | 36.84% | 9.22% |
| | **Total** | **124** | **82** | **206** | 60.19% | 39.81% | 100.00% |
| | | | | | | | |
| Fully group-disjoint | Train | 41 | 20 | 61 | 67.21% | 32.79% | 54.46% |
| | Validation | 11 | 13 | 24 | 45.83% | 54.17% | 21.43% |
| | Test | 15 | 12 | 27 | 55.56% | 44.44% | 24.11% |
| | **Total** | **67** | **45** | **112** | 59.82% | 40.18% | 100.00% |

## 2.2 Graph representation

Each finite group is represented by its undirected Cayley graph, which serves as the input to the neural network. Given a finite group and a fixed symmetric generating set, each group element is represented as a node, while an edge connects two nodes whenever one is obtained from the other by multiplication by a generator. This representation preserves the algebraic structure of the group in a graph form that can be processed by the proposed GNN framework. The constructed Cayley graph is then converted into a graph object consisting of its nodes, edges, and the associated connectivity information required by the shared GNN encoder. The same graph construction procedure is applied to both the subgroup candidate $H$ and the ambient group $G$, ensuring a consistent representation before they are processed by the Siamese GNN. In addition to the predefined generating sets used in the main experiments, alternative generating sets are considered in a separate experiment to evaluate the sensitivity of the model to different Cayley graph representations of the same groups.

## 2.3 Siamese GNN

The proposed model consists of a Siamese GNN designed to compare two input groups, namely the subgroup candidate $H$ and the ambient group $G$. Each input group is represented by its undirected Cayley graph and processed independently by a shared GNN encoder. The two branches have identical architectures and share the same parameters, ensuring that both graphs are embedded into a common latent feature space. In addition to the learned graph representations, the proposed framework incorporates algebraic features computed directly from the input groups. The following subsections describe each component of the proposed Siamese GNN.

### 2.3.1 GNN Encoder

The shared GNN encoder receives the undirected Cayley graph of each input group and maps it to a graph-level embedding that captures its structural properties. Since both branches share the same network parameters, the resulting graph embeddings are learned in a common latent feature space, enabling meaningful comparison between the two input groups.

The GNN encoder operates through a sequence of message-passing layers. At each layer, every node aggregates information from its immediate neighbors according to the connectivity of the Cayley graph. The aggregated neighborhood information is then combined with the node's current representation and processed by a learnable neural layer to produce an updated node embedding. After several message-passing iterations, the final node embeddings are aggregated using a graph pooling operation to obtain graph-level embeddings for the two input groups, denoted by $z_H$ and $z_G$.

### 2.3.2 Algebraic Features

In addition to the learned graph embeddings, the proposed framework incorporates four algebraic feature types computed directly from the input groups. These features provide complementary information that may not be fully captured by the graph embeddings alone. Depending on the experimental setting, the algebraic feature vector includes the ratio of the group orders, $|G|/|H|$, the remainder, $|G| \ mod \ |H|$, the sizes of the centers of $H$ and $G$, and the sizes of the derived subgroups of $H$ and $G$ [14]. These features complement the learned graph embeddings by explicitly incorporating algebraic information into the prediction process.

### 2.3.3 Joint Feature Vector and Classifier

The prediction module combines the two complementary sources of information produced by the proposed framework: graph embeddings learned by the shared GNN encoder and algebraic features computed directly from the input groups. These components are combined through the selected graph-derived and algebraic features to form a joint feature vector describing the relationship between the two groups. The graph-derived features capture relationships between the learned representations of the two groups, whereas the algebraic features provide explicit group-theoretic information that complements the learned embeddings. The different feature configurations are evaluated experimentally to determine which combination provides the most informative representation for subgroup prediction. Depending on the experimental configuration, the joint feature vector may include one or more of the features listed in Table 2.

**Table 2.** Features used to construct the joint feature vector.

| Feature | Abbrev. | Description |
|---|---|---|
| $z_H$ | $z_H$ | Graph-level embedding of group $H$ |
| $z_G$ | $z_G$ | Graph-level embedding of group $G$ |
| Diff | D | Element-wise absolute difference between the graph embeddings, $\|z_H - z_G\|$ |
| Product | P | Element-wise product of the graph embeddings, $z_H \odot z_G$ |
| Ratio | Q | Ratio of the group orders $\|G\|/\|\mathrm{H}\|$ |
| Remainder | R | Remainder of the group orders $\|G\| \, mod \, \|H\|$ |
| Center | C | Size of the center of the group |
| Derived | V | Size of the derived subgroup of the group |

## 2.4 Model training

The proposed Siamese GNN was trained as a supervised binary classifier. Each training sample consisted of an ordered pair of finite groups, their corresponding Cayley graph representations, and a binary label indicating whether $H$ is isomorphic to a subgroup of $G$. A label of 1 indicates that $H$ is isomorphic to a subgroup of $G$, whereas a label of 0 indicates that no such subgroup relation exists.

The model parameters were optimized using the Adam optimizer with a learning rate of 0.005. To account for class imbalance in the training data, a class-weighted binary cross-entropy loss with logits (BCEWithLogitsLoss) was used. The positive-class weight was defined as the ratio between the number of negative and positive training samples, $n0/n1$, thereby assigning greater weight to the minority class when the classes were imbalanced. For each training pair, the model produced a single output logit, which was compared with the corresponding binary target label. Gradients were computed by backpropagation, and the model parameters were updated after each individual training sample, corresponding to an effective batch size of one. Each model was trained for 100 epochs. During every epoch, all training samples were processed sequentially in their existing order, and the accumulated training loss was monitored throughout the optimization process.

To investigate the effect of network capacity, five GNN architectures were evaluated: (2-8-8), (2-16-16), (2-32-32), (2-64-64), and (2-128-128). In each architecture, the first number denotes the input node-feature dimension, while the remaining two numbers denote the hidden dimensions of the first and second GNN layers, respectively. The hidden layer of the classifier uses the same hidden dimension as the GNN layers.

Different feature configurations, described in Section 2.3, were evaluated as part of an ablation study to assess the contribution of graph-derived and algebraic features to subgroup prediction. Each trained model configuration, defined by its GNN architecture and feature configuration, was saved separately for subsequent evaluation.

For each data-split setup, the different GNN architectures and feature configurations were evaluated on the validation set using balanced accuracy (BA). The configuration achieving the highest validation BA was selected as the best-performing model and was subsequently evaluated on the corresponding test set. The test set was therefore not used for model selection.

### 2.5 Evaluation Metrics

For model evaluation, we report accuracy (Acc.), BA, precision (Prec.), recall (Rec.), F1-score (F1), and the area under the receiver operating characteristic curve (AUC). In addition, the numbers of true positives (TP), false negatives (FN), true negatives (TN), and false positives (FP) are reported to provide a complete summary of the classification outcomes.

## 3 Results

Three data-split setups are evaluated: Standard, Test-group-disjoint, and Fully group-disjoint, as described in Section 2.1. For each split setup and feature configuration, several GNN architectures are evaluated on the validation set, and the architecture achieving the highest BA is reported. The validation results are presented in Table 3. Within each data-split setup, the feature configurations are ordered by validation BA from highest to lowest, and the top-performing configuration is selected for subsequent evaluation on the corresponding test set.

**Table 3.** Validation results under different data split setups.

| Split setup | Features | Best architecture | BA |
|---|---|---|---|
| Standard split | Diff + Remainder + Center | 2-16-16 | 0.9333 |
| | Diff + Remainder | 2-8-8 | 0.9172 |
| | Diff + Remainder + Center + Derived | 2-64-64 | 0.9172 |
| | $z_H$ + $z_G$ + Remainder | 2-64-64 | 0.9005 |
| | Diff + Remainder + Derived | 2-16-16 | 0.9005 |
| | Remainder | 2-8-8 | 0.8500 |
| | $z_H$ + $z_G$ + Diff + Product | 2-8-8 | 0.8204 |
| | $z_H$ + $z_G$ + Diff | 2-8-8 | 0.8188 |
| | $z_H$ + $z_G$ + Diff + Product + Ratio | 2-16-16 | 0.7860 |
| | Derived | 2-16-16 | 0.7339 |
| | Derived + Center | 2-8-8 | 0.6833 |
| | Center | 2-8-8 | 0.4634 |
| | | | |
| Test-group-disjoint | Diff + Remainder | 2-128-128 | 0.9444 |
| | Diff + Remainder + Center | 2-64-64 | 0.9352 |
| | $z_H$ + $z_G$ + Remainder | 2-8-8 | 0.9167 |
| | Diff + Remainder + Derived | 2-16-16 | 0.9167 |
| | Diff + Remainder + Center + Derived | 2-64-64 | 0.9167 |
| | Remainder | 2-8-8 | 0.8796 |
| | $z_H$ + $z_G$ + Diff + Product | 2-8-8 | 0.8519 |
| | $z_H$ + $z_G$ + Diff | 2-32-32 | 0.8426 |
| | $z_H$ + $z_G$ + Diff + Product + Ratio | 2-16-16 | 0.8333 |
| | Derived + Center | 2-16-16 | 0.8148 |
| | Derived | 2-8-8 | 0.6852 |
| | Center | 2-32-32 | 0.4722 |
| | | | |
| Fully group-disjoint | Diff + Remainder + Center + Derived | 2-128-128 | 0.9231 |
| | Remainder | 2-8-8 | 0.9091 |
| | Diff + Remainder | 2-16-16 | 0.9091 |
| | $z_H$ + $z_G$ + Remainder | 2-16-16 | 0.9091 |
| | Diff + Remainder + Center | 2-128-128 | 0.9091 |
| | $z_H$ + $z_G$ + Diff + Product | 2-128-128 | 0.8636 |
| | $z_H$ + $z_G$ + Diff + Product + Ratio | 2-8-8 | 0.8252 |
| | $z_H$ + $z_G$ + Diff | 2-16-16 | 0.8252 |
| | Diff + Remainder + Derived | 2-8-8 | 0.8007 |
| | Derived | 2-8-8 | 0.7483 |
| | Derived + Center | 2-8-8 | 0.7028 |
| | Center | 2-64-64 | 0.5350 |

After selecting the best-performing model for each split setup based on the validation results in Table 3, the selected models are evaluated on their corresponding test sets. Three split setups are considered: standard, test-group-disjoint, and fully group-disjoint. The corresponding test-set results are presented in Table 4. For each setup, the table reports the selected feature configuration and GNN architecture together with the complete test-set performance. The reported measures include TP, FN, TN, FP, accuracy, BA, precision, recall, F1-score, and AUC.

**Table 4.** Performance of the best-performing model for each dataset split.

| Split | Arch. | Feat. | TP | FN | TN | FP | Acc. | BA | Prec. | Recall | F1 | AUC |
|---|---|---|---|---|---|---|---|---|---|---|---|---|
| S | (2-16-16) | D+R+C | 39 | 0 | 45 | 9 | 0.9032 | 0.9167 | 0.8125 | 1.0000 | 0.8966 | 0.9601 |
| T | (2-128-128) | D+R | 6 | 1 | 11 | 1 | 0.8947 | 0.8869 | 0.8571 | 0.8571 | 0.8571 | 0.9405 |
| F | (2-128-128) | D+R+C+V | 11 | 1 | 15 | 0 | 0.9630 | 0.9583 | 1.0000 | 0.9167 | 0.9565 | 0.9944 |

**Note:** Feature abbreviations are defined in Table 2. S = standard split; T = test-group-disjoint split; F = fully group-disjoint split. Acc. = accuracy; BA = balanced accuracy; Prec. = precision; AUC = area under the ROC curve.

For the error analysis, we consider the best-performing Siamese GNN configuration selected on the validation set under the standard split. As reported in Table 3, this model uses the 2-16-16 architecture with the D + R + C feature configuration. The model was evaluated on the corresponding standard-split test set, and the misclassified instances were examined individually. Table 5 reports these misclassified group pairs together with their true and predicted labels and prediction probabilities.

**Table 5.** Error analysis of the selected Siamese GNN on the standard-split test set.

| Group 1 | Group 2 | True | Pred. | Prob. (1) |
|---|---|---|---|---|
| C8 | S5 | 0 | 1 | 0.7749 |
| L2 | L3 | 0 | 1 | 0.9995 |
| C12 | S4 | 0 | 1 | 0.6941 |
| Q8 | S4 | 0 | 1 | 0.9239 |
| D10 | S5 | 0 | 1 | 0.8565 |
| C6 | S4 | 0 | 1 | 0.8745 |
| F5 | D10 | 0 | 1 | 0.9790 |
| C12 | S5 | 0 | 1 | 0.6747 |
| C4 | A4 | 0 | 1 | 0.9240 |

**Note:** Group-family abbreviations appearing in the table are as follows: C = cyclic; D = dihedral; S = symmetric; A = alternating; Q = quaternion; L = special linear; F = affine general linear.

To evaluate the stability of the model across different data partitions, five-fold stratified cross-validation is performed on the full dataset. The dataset is divided into five folds, and in each run four folds are used for training while the remaining fold is used for evaluation. This procedure is repeated until each fold has served once as the evaluation fold. Table 6 reports the mean BA and standard deviation obtained across the five folds.

**Table 6.** Five-fold cross-validation results.

| Arch. | Features | Rate | Epochs | Folds | Seed | Mean BA | Std. BA |
|---|---|---|---|---|---|---|---|
| (2-16-16) | D + R + C | 0.0050 | 100 | 5 | 42 | 0.9052 | 0.0370 |

The cross-validation results indicate that the model maintains consistent performance across different partitions of the dataset, supporting the stability of the proposed approach.

To evaluate the contribution of the GNN component, the proposed Siamese GNN is compared with two baseline configurations. The comparison uses the best-performing model selected on the validation set under the standard split, with architecture 2-16-16 and features D + R + C. First, the message-passing component is removed while the remaining feature-processing and classification components are retained, allowing the contribution of graph message passing to be examined directly. Since the embedding-difference feature D requires GNN-generated graph embeddings, it is not available when the GNN component is removed. Therefore, the No-GNN configuration uses only the corresponding algebraic features R + C. Second, a non-neural logistic regression model based on the ratio and remainder features is evaluated as a simple algebraic baseline. Table 7 compares the performance of these approaches on the standard-split test set using the same evaluation metrics.

**Table 7.** Model ablation and baseline comparison.

| Model | Message passing | Feat. | Acc. | BA | Prec. | Recall | F1 | AUC |
|---|---|---|---|---|---|---|---|---|
| GNN | Yes | D+R+C | 0.9032 | 0.9167 | 0.8125 | 1.0000 | 0.8966 | 0.9601 |
| No-GNN | No | R+C | 0.8710 | 0.8889 | 0.7647 | 1.0000 | 0.8667 | 0.8889 |
| Logistic-Reg | No | R+Q | 0.8710 | 0.8889 | 0.7647 | 1.0000 | 0.8667 | 0.9364 |

**Note:** Feature abbreviations are defined in Table 2. Acc. = accuracy; BA = balanced accuracy; Prec. = precision; AUC = area under the ROC curve.

The results in Table 7 show that the full Siamese GNN achieves the best overall performance among the considered configurations. The comparison indicates that incorporating graph-based message passing provides additional predictive information beyond the algebraic features alone, resulting in improved performance over the non-GNN baselines.

To examine whether the model's performance depends on the particular generating sets used to construct the Cayley graphs, an additional experiment is performed using alternative, independently specified generating sets. Two generating-set configurations are considered for each group pair $(H, G)$. In the Original configuration, both $H$ and $G$ are represented using the predefined generating sets used in the main experiments. To construct the Alternative configuration, an additional generating-set specification is defined for each group family, providing a second way of generating the corresponding Cayley graph. For each group pair, the alternative generating sets are used to construct the Cayley graphs of $H$ and $G$ independently. Consequently, the representation of $H$ is not obtained from or inherited from the representation of $G$. The model is then trained and evaluated using these independently constructed Cayley graphs, following the same general learning procedure as in the main experiment. The corresponding results are reported in Table 8.

**Table 8.** Performance under different Cayley graph generating sets.

| Generating set | Acc. | BA | AUC |
|---|---|---|---|
| Regular | 0.9032 | 0.9167 | 0.9601 |
| Non-inherited | 0.8925 | 0.9003 | 0.9601 |

The results in Table 8 show that the model maintains comparable performance when alternative generating sets are used. This indicates that the classification performance is reasonably robust to the choice of Cayley graph representation and does not strongly depend on the representation of H being inherited from that of G.

For comparison with a classical computational approach, the best-performing Siamese GNN configuration selected on the validation set under the standard split was compared with GAP. As reported in Tables 3 and 4, this model uses the 2-16-16 architecture with the D + R + C feature configuration. For a direct and consistent comparison, both methods were evaluated on exactly the same standard-split test instances. Their predictive performance and computational runtime are reported in Table 9.

**Table 9.** Comparison of the Siamese GNN and GAP on the standard-split test set.

| Method | Acc. | BA | Runtime (s) |
|---|---|---|---|
| Siamese GNN | 0.9032 | 0.9167 | 0.127 |
| GAP | 1.000 | 1.000 | 0.297 |

The comparison shows that GAP achieves perfect classification on the considered test set, whereas the Siamese GNN provides lower predictive performance but requires less computation time for inference. These results highlight the expected advantage of the exact computational approach in terms of correctness, while demonstrating that the learned model can provide predictions with comparatively low inference cost.

## 4 Discussion

This section discusses the main findings of the experimental evaluation and examines the contributions of the proposed framework. In particular, it considers the effectiveness of the Cayley graph representation, the role of the Siamese architecture, the contribution of the algebraic features, and the insights obtained from the ablation study. In addition, the discussion considers the generalization of the model across different data splits, its robustness to different generating sets, its comparison with classical computational methods, and the observed misclassification patterns.

Overall performance. The proposed Siamese GNN was evaluated on a dataset comprising 308 finite-group pairs drawn from 11 group families. The proposed Siamese GNN achieved a test BA of 91.67% on the independent standard-split test set, demonstrating its ability to learn subgroup relations directly from graph representations of finite groups. The evaluation was further extended to the Test-group-disjoint and Fully group-disjoint settings, where the model achieved balanced accuracies of 0.8869 and 0.9583, respectively. These experiments provide a stricter evaluation of the model by reducing or eliminating overlap of groups between the different data partitions. Five-fold cross-validation further resulted in a mean BA of 0.9052 with a standard deviation of 0.0370, indicating relatively stable performance across different data partitions.

Effectiveness of Cayley graphs. The experimental results indicate that the combination of Cayley graph representations and the proposed Siamese GNN provides an effective framework for learning subgroup relations between finite groups. The GNN encoder is able to learn informative structural representations from the Cayley graphs, while the Siamese architecture enables meaningful comparison between pairs of groups. Since the Cayley graph depends on the chosen generating set, different generating sets may lead to different graph representations and consequently influence the learned embeddings and predictive performance. To examine this dependence directly, the model was also evaluated using alternative generating sets, in which the Cayley graph of H was constructed independently of the Cayley graph of G. The model achieved a BA of 0.9003 under this setting, compared with 0.9167 under the original configuration. This result provides an indication of the extent to which the learned predictions are affected by the particular Cayley graph presentation.

Role of the Siamese architecture. The Siamese architecture is well suited to the subgroup prediction task because it is specifically designed to learn relationships between pairs of inputs. By processing both groups with a shared encoder, the proposed framework learns graph embeddings in a common latent feature space, enabling consistent comparison of their structural properties. Furthermore, the modular separation between the encoder and the classifier provides flexibility, allowing each component to be modified or replaced independently without changing the overall framework. The contribution of GNN message passing was further examined by comparing the complete Siamese GNN with a configuration in which message passing was removed. The corresponding balanced accuracies were 0.9167 and 0.8889, respectively. The model was also compared with a non-neural logistic regression baseline based on algebraic features, which achieved a BA of 0.8889. These comparisons provide a direct assessment of the contribution of graph-based representation learning and message passing to the prediction task.

Contribution of algebraic features. The experimental results demonstrate that the algebraic features complement the graph embeddings learned by the GNN encoder. While the graph embeddings capture structural information from the Cayley graphs, the algebraic features explicitly encode fundamental properties of the input groups that may not be directly represented in the learned embeddings. The experimental results indicate that combining graph-based and algebraic information can improve subgroup prediction compared with the non-GNN baselines considered in this study. Additional algebraic features were also evaluated to determine whether richer group-theoretic information could further improve predictive performance. The results in Table 3 show that the best-performing configuration under the standard split combined the embedding-difference feature with the remainder and center features, while the other feature configurations considered did not achieve higher validation performance.

Ablation study. The ablation study demonstrates the contribution of the different components of the proposed framework. Comparing different combinations of graph-derived and algebraic features shows that these components provide complementary information. The highest validation BA was achieved by combining the embedding-difference feature with the remainder and center features, indicating that this combination provides a more informative representation for subgroup prediction than the other feature configurations considered. Overall, the results demonstrate the benefit of integrating learned graph representations with domain-specific algebraic information. One possible explanation for the usefulness of the remainder feature is that divisibility of group orders provides direct algebraic information relevant to subgroup relations: by Lagrange's theorem, the order of a subgroup must divide the order of the ambient group. The remainder therefore provides information about whether this necessary condition is satisfied. However, divisibility alone is not sufficient to determine a subgroup relation, which helps explain why combining this algebraic information with learned graph representations can be beneficial.

Comparison with classical computation. The proposed learning-based approach was also compared directly with GAP on the same test instances. The Siamese GNN achieved an accuracy of 0.9032 and a BA of 0.9167, with a runtime of 0.127 s, whereas GAP achieved an accuracy of 1.0000 and a BA of 1.0000, with a runtime of 0.297 s. This comparison is not intended to suggest that the probabilistic model should replace exact computational group-theoretic methods. Rather, it provides a direct quantitative comparison between the learning-based approach and an established exact computational approach.

Error analysis. The misclassified test instances were also examined to identify possible patterns in the model's errors. Among the 93 test instances, 9 were classified incorrectly. The errors consisted of 9 FPs and 0 FNs. This indicates that, for the misclassified instances in the standard-split test set, the model's errors were exclusively cases in which a non-subgroup relation was incorrectly predicted as a subgroup relation, and provides additional insight into cases in which the learned model has difficulty distinguishing subgroup relations.

## 5 Limitations

The proposed framework has several limitations that should be acknowledged. First, the experimental evaluation was conducted on a finite dataset of groups, and the model's performance may depend on the size and diversity of the available data. Second, the Cayley graph representation depends on the choice of generating set, and different generating sets may produce different graph structures and, consequently, different learned representations. Third, the proposed model is a probabilistic predictor and therefore does not provide the exact guarantees offered by classical computational group-theoretic methods. Finally, the proposed framework represents only one possible architectural design. Different GNN encoders, classifier architectures, or richer algebraic feature sets may lead to different predictive performance. Investigating these aspects remains an important direction for future research.

## 6 Future work

Future work may explore several extensions of the proposed framework. Since the Siamese architecture is modular, the shared GNN encoder may be replaced by alternative neural network models, or its internal architecture may be modified by employing different GNN designs. Likewise, different classifier architectures and richer algebraic feature sets may be investigated. In addition, further expansion to more diverse and structurally complex group families, as well as a broader range of Cayley graph presentations, may be investigated. These directions may further improve the robustness, generalization, and predictive performance of the proposed framework.

## 7 Conclusions

The proposed framework illustrates the potential of combining computational group theory with geometric deep learning to address algebraic learning problems. While this work focuses on subgroup prediction, the same general approach may be extended to other relational problems involving finite groups and related algebraic structures. More broadly, the results suggest that graph-based neural networks, when combined with suitable algebraic representations and domain-specific features, provide a promising framework for learning complex mathematical relationships directly from structured algebraic data.